\documentclass[11pt]{article}
\usepackage[preprint]{acl}
\usepackage{times}
\usepackage{latexsym}
\usepackage[T1]{fontenc}
\usepackage{svg}
\usepackage[utf8]{inputenc}
\usepackage{microtype}
\usepackage{inconsolata}
\usepackage{array}
\usepackage{graphicx}
\usepackage{multirow}
\usepackage{booktabs}
\usepackage{ragged2e}
\usepackage{vcell}
\usepackage{color}
\usepackage{siunitx} 
\usepackage{subfig}
\usepackage[normalem]{ulem}
\usepackage{diagbox}

\usepackage{graphicx}

\usepackage{algorithm}
\usepackage{algorithmic}
\usepackage{tabularx}
\usepackage{natbib}

\title{Correct after Answer: Enhancing Multi-Span Question Answering with Post-Processing Method}

\author{
    \textbf{Jiayi Lin \textsuperscript{1,2}}, 
    \textbf{Chenyang Zhang \textsuperscript{1,2}}, 
    \textbf{Haibo Tong \textsuperscript{1,2}}, 
    \textbf{Dongyu Zhang \textsuperscript{1,2}} 
    \\
    \textbf{Qingqing Hong \textsuperscript{1,2}}, 
    \textbf{Bingxuan Hou \textsuperscript{1,2}}, 
    \textbf{Junli Wang \textsuperscript{1,2}\thanks{Corresponding author. This work was supported by the National Key Research and Development Program of China under Grant 2023YFB3002201.}} 
    \\
    \textsuperscript{1} Key Laboratory of Embedded System and Service Computing (Tongji University),\\ Ministry of Education, Shanghai 201804, China.\\
    \textsuperscript{2} National (Province-Ministry Joint) Collaborative Innovation Center \\for Financial Network Security, Tongji University, Shanghai 201804, China.\\
    \texttt{\{2331908, inkzhangcy, 2151130, yidu\}@tongji.edu.cn}\\\texttt{\{2332012, 2052643, junliwang\}@tongji.edu.cn}\\
}

\begin{document}
\maketitle
\begin{abstract}

Multi-Span Question Answering (MSQA) requires models to extract one or multiple answer spans from a given context to answer a question. Prior work mainly focuses on designing specific methods or applying heuristic strategies to encourage models to predict more correct predictions. However, these models are trained on gold answers and fail to consider the incorrect predictions. Through a statistical analysis, we observe that models with stronger abilities do not predict less incorrect predictions compared with other models. In this work, we propose \textbf{A}nswering-\textbf{C}lassifying-\textbf{C}orrecting (ACC) framework, which employs a post-processing strategy to handle incorrect predictions. Specifically, the ACC framework first introduces a \textbf{classifier} to classify the predictions into three types and exclude "wrong predictions", then introduces a \textbf{corrector} to modify "partially correct predictions". Experiments on several MSQA datasets show that ACC framework significantly improves the Exact Match (EM) scores, and further analysis demostrates that ACC framework efficiently reduces the number of incorrect predictions, improving the quality of predictions.\footnote[1]{Our code and data are available at \url{https://github.com/TongjiNLP/ACC}.}

\end{abstract}

\section{Introduction}

\begin{figure}[t]
  \includegraphics[width=\columnwidth]{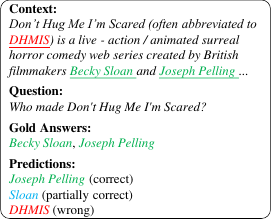}
  \caption{An example of MSQA. This question has two gold answers: "Becky Sloan" and "Joseph Pelling". "Joseph Pelling" is a correct prediction, "Sloan" is a partially correct prediction and "DHMIS" is a wrong prediction. Best read in colors.}
  \label{fig:example_of_MSQA}
\end{figure}

Machine Reading Comprehension (MRC) requires models to answer a question based on a given context \cite{rajpurkar-2018-squad, kwiatkowski-2019-nq, lai-2017-race}. In a real-world scenario, a single question typically corresponds to multiple answers. To this end, Multi-Span Question Answering (MSQA) has been proposed \cite{ju-2022-cmqa,li-2022-msqa,yue-2023-mamrc}. Different from the traditional Single-Span Question Answering (SSQA) task, the goal of MSQA is to extract one or multiple non-overlapped spans from the given context.
Taking Figure~\ref{fig:example_of_MSQA} as an instance, the question "Who made Don't Hug Me I'm Scared?" has two answers: "Becky Sloan" and "Joseph Pelling".

Recent MSQA work integrates various approaches.
~\citet{yang-2021-musst, hu-2019-mtmsn} incorporate heuristic strategies based on traditional pointer models \cite{vinyals-2015-pointer} to extract multiple answers;
~\citet{segal-2020-tase,li-2022-msqa} convert MSQA task into a sequence-tagging task to mark answers;
~\citet{huang-2023-spanqualifier, zhang-2023-css} enumerate all candidate answers and select the final answers with a learnable threshold,
and ~\citet{huang-2023-fbprompt,zhang-2023-howmany} utilize Large Language Models (LLMs) to handle MSQA tasks with few shot prompts.

Prior work mainly focus on specific methods or heuristic strategies for more correct predictions.
However, these models are trained on gold answers, and fail to consider the incorrect predictions.
To further investigate the incorrect predictions, we categorize predictions into "correct predictions", "partially correct predictions" and "wrong predictions" based on whether they should be modified or excluded.
Then we conduct a statistical analysis on several MSQA models (details in Section~\ref{subsec:analysis_of_msqa}), and observe that stronger MSQA models~\footnote{"stronger" indicates higher MSQA performance, which can be demonstrated using EM scores.} do not predict less incorrect predictions compared with other models.
Consequently, performance of the MSQA models can be further improved on the basis of reducing incorrect predictions.

In this work, we propose \textbf{A}nswer-\textbf{C}lassify-\textbf{C}orrect (ACC) framework, which employs a post-processing strategy to handle with incorrect predictions.
The ACC framework simulates humans strategy in realword examinations: listing candidate answers, reviewing and modifying.
Specifically, we design the \textbf{classifier} to categorize candidate answers into "correct predictions", "partially correct predictions" or "wrong predictions",
then we design the \textbf{corrector} to modify "partially correct predictions", finally we exclude "wrong predictions" and obtain final predictions. To train the classifier and the corrector, we also apply an automatic annotation approach which samples incorrect predictions from the training datasets and constructs the silver-labeled datasets.

We conduct experiments on four MSQA datasets. Experiment results show that the ACC framework significantly improves the performance.
After applying the ACC framework, the EM F1 score increases from 69.05\% to 72.26\% for Tagger-RoBERTa \cite{li-2022-msqa} and from 65.57\% to 76.31\% for BART-base \cite{lewis-2020-bart} on the MultiSpanQA dataset \cite{li-2022-msqa}. Further analysis on the predictions indicates that the ACC framework effectively reduces the number of incorrect predictions and obtains more correct predictions, enhancing the qualities of predictions.
In addition, a pilot study with GPT-3.5 \footnote{\url{https://platform.openai.com/}.} is conducted, exhibiting extensive application of ACC framework for LLMs in a Chain-of-Thought (CoT) manner~\cite{wang-2022-fscot, kojima-2022-zscot}.

Our contributions are summarized as follows:

\begin{itemize}
  \item We develop a three-fold taxonomy for the MSQA predictions based on whether a prediction should be modified or excluded. Then, we conduct a statistical analysis, revealing distributions over the three categories.
  \item Inspired by humans' strategies, we propose the ACC framework, which includes a classifier to exclude incorrect predictions and includes a corrector to modify imperfect predictions. We also design an automatical annotation approach to sample incorrect predictions and construct silver-labeled datasets.
  \item We conduct experiments and analysis on several MSQA datasets. Results show that the ACC framework significantly enhances the quality of the MSQA predictions.
\end{itemize}

\begin{figure*}[t]
  \includegraphics[width=1\linewidth]{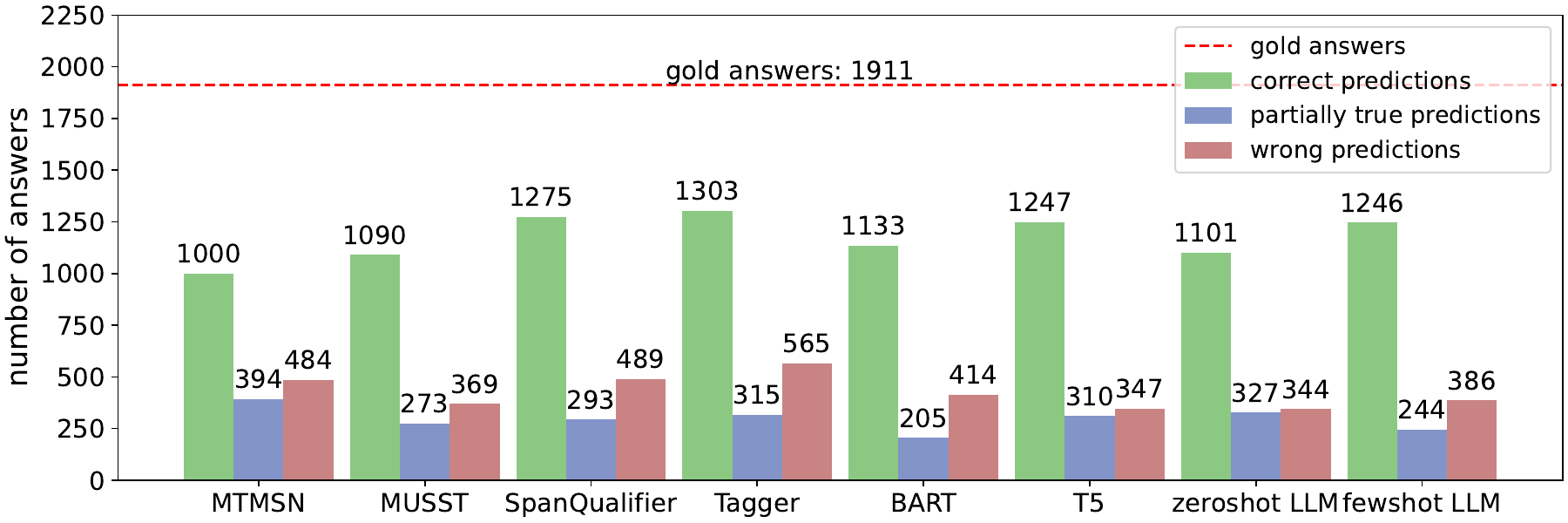}\hfill
  \caption{The distribution of correct predictions, partially correct predictions and wrong predictions on the validation set of MultiSpanQA. The validation set of MultiSpanQA contains 653 questions with 1,911 gold answers.}
  \label{fig:distribution_of_predictions}
\end{figure*}

\section{Taxonomy of MSQA Predictions}

\subsection{Formalization}

Given a question $Q$ and its corresponding context $C$, the goal of MSQA is to train a model $M$ to extract a set of $m$ answer spans $P=\{p_1,p_2,...,p_m\}$ from the given context, shown as Eq.~\ref{eq:msqa}.

\begin{equation}
\label{eq:msqa}
  P=M(C,Q)
\end{equation}

\subsection{Taxonomy}
\label{subsec:types_of_answers}

Intuitively, the predictions can be categorized as correct or incorrect predictions. However, some of incorrect predictions should be modified while others should be excluded.
For instances in Figure~\ref{fig:example_of_MSQA}, "Sloan" and "DHMIS" are both incorrect predictions.
However, "Sloan" is similar to the gold answer "Bercy Sloan" but "DHMIS" is totally wrong.
Therefore, we further categorize incorrect predictions into "partially correct predictions" and "wrong predictions".

Based on above analysis, we category the prediction $p_i \in P$ into one of the following three types: "correct prediction", "partially correct prediction" and "wrong prediction".

\paragraph{Correct prediction.} If the prediction $p_i$ is one of the gold answers, which means $p_i \in A$, $p_i$ is regarded as a correct prediction.

\paragraph{Partially correct prediction.} We utilize Word Overlap (WO) and BERTScore (BS) \cite{zhang-2020-bertscore} to define partially correct predictions. Word Overlap considers the overlap between two spans in word level, while BERTScore computes semantic similarity of two spans in the manner of cosine similarity.
Details of Word Overlap and BERTScore are shown in Appendix~\ref{sec:wo_bs}.

For a prediction $p_i$, if there exists $a_j \in A$ which satisfies $WO(p_i,a_j) \ge \alpha$ and $BS(p_i,a_j) \ge \beta$, where $\alpha$ and $\beta$ are hyper-parameters, the $p_i$ is regarded as the partially correct prediction.
  
\paragraph{Wrong prediction.} If $p_i$ is not a correct prediction or a partially correct prediction, the $p_i$ is regarded as wrong prediction.

Figure~\ref{fig:example_of_MSQA} shows an example containing these three types of predictions. The gold answers are "Becky Sloan" and "Joseph Pelling". Among the three predictions, "Joseph Pelling" is a correct prediction; "Sloan" is a partially correct prediction and "DHMIS" is a wrong prediction.

\begin{figure*}[t]
  \includegraphics[width=1\linewidth]{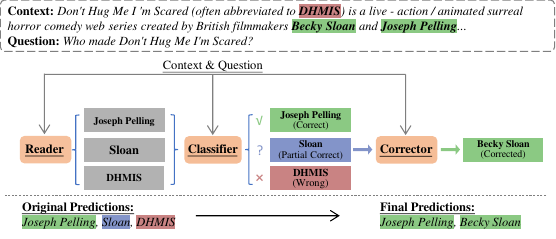}\hfill
  \caption{The overall architecture of our proposed ACC framework.}
  \label{fig:architecture_of_ACC}
\end{figure*}

\subsection{Distribution of MSQA Predictions}
\label{subsec:analysis_of_msqa}

Based on the taxonomy, we conduct a statistical analysis on the validation set of MultiSpanQA~\cite{li-2022-msqa}. We select four discriminative models: MTMSN \cite{hu-2019-mtmsn}, MUSST \cite{yang-2021-musst}, Tagger \cite{li-2022-msqa} and SpanQualifier \cite{huang-2023-spanqualifier}, and three generative models: BART \cite{lewis-2020-bart}, T5 \cite{raffel-2020-t5} and GPT-3.5 (zero-shot and few-shot).
Details of these models are shown in Appendix~\ref{subsec:details_of_baselines}.

The statistical results are shown in Figure~\ref{fig:distribution_of_predictions}. We observed that models with better performance (shown in Table \ref{table:main_result_msqa}) on the validation set predict more correct predictions as well as more wrong predictions.
For example, for discriminative models, Tagger predicts 1,303 correct predictions but also predict 565 wrong predictions, while MTMSN predicts 1,000 correct predictions and 484 wrong predictions; Similarly, after adding few-shot demonstrations, the LLM generates more correct as well as wrong predictions compared with zero-shot setting. Therefore, we believe that the post-processing method can effectively enhance the quality of predictions by reducing the number of incorrect predictions, resulting in better performance.

\section{Method}

In this section, we describe the ACC framework, which is designed to handle with partially correct predictions and wrong predictions. The architecture of the ACC framework is shown in Figure~\ref{fig:architecture_of_ACC}.

Similar to the humans' strategies, the post-processing procedure of the ACC framework consists of three steps: The first step is \textbf{answering}, where we employ a \textbf{reader} to obtain initial predictions $P$; The second step is \textbf{classifying}, where we employ a \textbf{classifier} to categorize each prediction $p_i$ into one of the three classes: correct prediction, partially correct prediction and wrong prediction; The last step is \textbf{correcting}, where we employ a \textbf{corrector} to modify the partially correct predictions. We reserve correct predictions predicted by the classifier and the modified predictions from the corrector as the final predictions.

Next, we will provide more details of the reader, the classifier and the corrector. We will also introduce an automatic annotation approach which samples incorrect predictions and constructs training data for the classifier and the corrector.

\subsection{Reader}
\label{subsec:ans_model}

The main function of the reader is to extract several text spans from context based on a given question. This process can be described as:

\begin{equation}
  P = Reader(Q,C)
\end{equation}

\subsection{Classifier}
\label{subsec:cls_model}

The predictions of the reader may include partially correct predictions or wrong predictions (mentioned in Section~\ref{subsec:types_of_answers}). To this end, we design the classifier to classify them and exclude wrong predictions. Given the candidate predictions $P$, the classifier splits them into correct predictions $P_c$, partially correct predictions $P_p$ and wrong predictions $P_w$. This process can be described as:

\begin{equation}
  P_c,P_p,P_w=Classifier(P,Q,C)
\end{equation}

Specifically, the classifier consists of a transformer \cite{vaswani-2017-attention} encoder and a classification head. The classification head includes an MLP layer to obtain probability of each class. Inspired by \citet{zhu-2022-duma}, we also add a cross-attention layer in the classification head. The cross-attention layer calculates the attention scores between the question and the context to enhance the representations of them.

\subsection{Corrector}
\label{subsec:cor_model}

The classifier is able to exclude wrong predictions, however, there may still contain partially correct predictions which are imperfect and should be modified. Hence, we design the corrector to modify those partially correct predictions. This process can be described as:

\begin{equation}
  \hat{P_p}=Corrector(P_p,Q,C)
\end{equation}
where $\hat{P_p}$ are the predictions modified by corrector.

We adopt traditional pointer model~\cite{vinyals-2015-pointer} to predict the start and end probabilities $st$ and $ed$. During the inference stage, for the text span starting at $i$-th token and ending at $j$-th token, we calculate its confidence score $score_{ij}=st_i + ed_j$ and obtain the best index pair $(i,j)$ which maximizes $score_{ij}$, then extract its corresponding span as the modified prediction. 

The final outputs of the ACC framework $\tilde{P}$ consist of the correct predictions $P_c$ predicted by the classifier and the modified predictions $\hat{P_p}$ from the corrector, described as:

\begin{equation}
  \tilde{P}=P_c \cup \hat{P_p}
\end{equation}

\subsection{Data Annotations}
\label{subsec:data_annotations}

To train the classifier and the corrector, we need both correct predictions and incorrect predictions. However, most MSQA datasets do not contain incorrect predictions. Inspired by \citet{gangi-2020-answercorrect}, we adopt an automatical sampling method similar to K-fold cross-validation, to collect incorrect predictions from the MSQA datasets and construct our silver-labeled datasets.

First, we randomly divide the training data $D$ into $K$ equal subsets: $D_1,D_2,...,D_K$. We perform $K$ iterations, in the $i$-th iteration we initialize a MSQA model $M$ (i.e. reader mentioned in Section~\ref{subsec:ans_model}) and train it with all training data except $D_i$, then sampling the predictions of $D_i$ with $M$. After $K$ iterations, we utilize the gold answers from training dataset $D$ to annotate all predictions, and construct the silver-labeled dataset~\footnote{More details are shown in Appendix~\ref{subsec:implementation_details}.}.

\begin{table*}
\small
  \centering
  \resizebox{\linewidth}{!}{
\begin{tabular}{l|ccc|ccc|ccc|ccc} 
\hline
\multirow{2}{*}{} & \multicolumn{3}{c|}{\textbf{MultiSpanQA}} & \multicolumn{3}{c|}{\textbf{MultiSpanQA-Expand}} & \multicolumn{3}{c|}{\textbf{MAMRC}} & \multicolumn{3}{c}{\textbf{MAMRC-Multi}} \\
 & EM P & EM R & EM F1 & EM P & EM R & EM F1 & EM P & EM R & EM F1 & EM P & EM R & EM F1 \\ 
\hline
\multicolumn{13}{c}{Discriminative Models (BERT-base)} \\ 
\hline
MTMSN & 51.76 & 41.69 & 46.18 & 60.88 & 51.46 & 55.78 & 72.65 & 77.41 & 74.96 & 71.50 & 76.71 & 74.01 \\
+ACC & \textbf{67.75} & \textbf{49.52} & \textbf{57.22} & \textbf{67.77} & \textbf{54.91} & \textbf{60.66} & \textbf{81.60} & 77.40 & \textbf{79.44} & \textbf{85.55} & \textbf{79.32} & \textbf{82.32} \\ 
\hline
MUSST & 61.44 & 53.74 & 57.33 & 67.48 & 59.71 & 63.36 & 76.28 & 79.00 & 77.62 & 75.68 & 78.12 & 76.88 \\
+ACC & \textbf{68.84} & \textbf{54.39} & \textbf{60.76} & \textbf{69.62} & \textbf{60.05} & \textbf{64.48} & \textbf{81.94} & 77.10 & \textbf{79.45} & \textbf{85.87} & \textbf{78.38} & \textbf{81.95} \\ 
\hline
Tagger & 56.66 & 65.46 & 60.74 & 52.81 & 55.92 & 54.30 & 77.15 & 81.83 & 79.42 & 74.71 & 76.74 & 75.70 \\
+ACC & \textbf{68.52} & \textbf{67.05} & \textbf{67.78} & \textbf{62.74} & \textbf{58.83} & \textbf{60.71} & \textbf{82.56} & 79.67 & \textbf{81.10} & \textbf{85.80} & \textbf{77.58} & \textbf{81.48} \\ 
\hline
SpanQualifier & 67.99 & 69.44 & 68.70 & 62.83 & 67.88 & 65.25 & 77.51 & 84.51 & 80.86 & 76.10 & 85.39 & 80.47 \\
+ACC & \textbf{72.04} & 67.82 & \textbf{69.86} & \textbf{65.78} & 67.13 & \textbf{66.45} & \textbf{82.40} & 80.76 & \textbf{81.57} & \textbf{85.67} & 83.37 & \textbf{84.51} \\ 
\hline
\multicolumn{13}{c}{Discriminative Models (RoBERTa-base)} \\ 
\hline
MTMSN & 59.86 & 49.97 & 54.47 & 63.39 & 56.00 & 59.47 & 73.94 & 78.36 & 76.08 & 71.69 & 77.47 & 74.46 \\
+ACC & \textbf{71.75} & \textbf{55.87} & \textbf{62.82} & \textbf{68.95} & \textbf{58.81} & \textbf{63.48} & \textbf{81.84} & 77.70 & \textbf{79.72} & \textbf{85.13} & \textbf{79.82} & \textbf{82.39} \\ 
\hline
MUSST & 69.82 & 61.94 & 65.64 & 69.29 & 63.16 & 66.08 & 78.01 & 79.71 & 78.85 & 76.69 & 77.16 & 76.92 \\
+ACC & \textbf{73.07} & 61.78 & \textbf{66.96} & \textbf{70.54} & 62.60 & \textbf{66.33} & \textbf{82.75} & 77.57 & \textbf{80.08} & \textbf{86.10} & \textbf{77.48} & \textbf{81.56} \\ 
\hline
Tagger & 66.22 & 72.14 & 69.05 & 64.35 & 65.66 & 64.99 & 79.47 & 83.59 & 81.48 & 75.85 & 78.19 & 77.00 \\
+ACC & \textbf{72.39} & 72.12 & \textbf{72.26} & \textbf{68.70} & \textbf{66.21} & \textbf{67.43} & \textbf{83.62} & 81.80 & \textbf{82.70} & \textbf{85.77} & \textbf{78.36} & \textbf{81.90} \\ 
\hline
SpanQualifier & 70.40 & 72.82 & 71.58 & 64.65 & 69.65 & 66.99 & 83.40 & 80.83 & 82.10 & 75.63 & 85.77 & 80.37 \\
+ACC & \textbf{73.69} & 71.32 & \textbf{72.47} & \textbf{67.68} & 68.53 & \textbf{68.09} & 82.83 & 81.88 & \textbf{82.35} & \textbf{85.14} & 83.77 & \textbf{84.45} \\ 
\hline
\multicolumn{13}{c}{Generative Models} \\ 
\hline
BART-base & 69.10 & 62.38 & 65.57 & 60.42 & 55.95 & 58.10 & 77.53 & 74.33 & 75.89 & 75.96 & 73.21 & 74.56 \\
+ACC & \textbf{73.90} & 61.80 & \textbf{67.31} & \textbf{63.68} & 55.70 & \textbf{59.43} & \textbf{80.47} & 72.47 & \textbf{76.26} & \textbf{81.26} & 71.22 & \textbf{75.91} \\ 
\hline
T5-base & 70.56 & 67.97 & 69.24 & 64.63 & 64.59 & 64.61 & 77.01 & 79.88 & 78.41 & 75.27 & 77.14 & 76.19 \\
+ACC & \textbf{73.93} & 66.20 & \textbf{69.85} & \textbf{67.43} & 63.32 & \textbf{65.31} & \textbf{80.79} & 77.43 & \textbf{79.07} & \textbf{80.65} & 74.73 & \textbf{77.58} \\ 
\hline
GPT3.5 (Zeroshot) & 64.83 & 60.86 & 62.78 & 39.60 & 53.68 & 45.58 & 45.45 & 57.34 & 50.71 & 57.00 & 63.27 & 59.97 \\
+ACC & \textbf{73.04} & \textbf{61.96} & \textbf{67.04} & \textbf{48.64} & \textbf{53.96} & \textbf{51.16} & \textbf{57.10} & \textbf{57.71} & \textbf{57.40} & \textbf{69.54} & \textbf{64.06} & \textbf{66.69} \\ 
\hline
GPT3.5 (Fewshot) & 68.94 & 68.18 & 68.56 & 42.44 & 58.13 & 49.06 & 58.42 & 73.79 & 65.21 & 65.38 & 76.68 & 70.58 \\
+ACC & \textbf{74.88} & 66.61 & \textbf{70.51} & \textbf{51.65} & 57.91 & \textbf{54.60} & \textbf{68.02} & 70.94 & \textbf{69.45} & \textbf{75.39} & 74.97 & \textbf{75.18} \\
\hline
\end{tabular}
  }
  \caption{\label{table:main_result_msqa} EM Scores on four MSQA datasets. "EM P" "EM R" "EM F1" refer to the precision, recall and F1 score under the EM metric, respectively. "Discriminative Models (BERT-base/RoBERTa-base)" refer to models that utilize BERT-base or RoBERTa-base as encoders. The results marked in \textbf{bold} means improvements after applying the ACC framework.}
\end{table*}

\section{Experiments}
\label{sec:experiments}

\subsection{Experimental Setup}

\paragraph{Datasets} Four MSQA datasets are integrated in experiments: MultiSpanQA~\cite{li-2022-msqa}, MultiSpanQA-Expand~\cite{li-2022-msqa}, MAMRC~\cite{yue-2023-mamrc} and an additional synthetic dataset MAMRC-Multi.
Details of these datasets are shown in Appendix~\ref{subsec:details_of_datasets}.

\paragraph{MSQA models} We set both discriminative models and generative models as readers. For discriminative models, we set MTMSN \cite{hu-2019-mtmsn}, MUSST \cite{yang-2021-musst}, Tagger \cite{li-2022-msqa} and SpanQualifier \cite{huang-2023-spanqualifier}; For generative models, we set BART \cite{lewis-2020-bart}, T5 \cite{raffel-2020-t5} and GPT-3.5.
Details of these models are shown in Appendix~\ref{subsec:details_of_baselines}.

\paragraph{Evaluation Metrics} We use \textbf{Exact Match Precision/Recall/F1 (EM P/R/F1)} \cite{li-2022-msqa} as the main metrics in our experiments. EM assign a score of 1 when a prediction fully matches one of the gold answers and 0 otherwise.

\paragraph{Implementation Details} For the classifier and corrector in the ACC framework, we use RoBERTa-base \cite{liu-2021-roberta} as encoder. For discriminative MSQA models, we use both BERT-base \cite{devlin-2019-bert} and RoBERTa-base as encoder. For the hyper parameters mentioned in Section~\ref{subsec:types_of_answers}, based on the average Word Overlap and BERTScore of the sampled data, we set $\alpha = 0.25$ and $\beta = 0.6$ to obtain balanced training data. See more training and inference details in Appendix~\ref{subsec:implementation_details}.

\subsection{Main Results}

Table~\ref{table:main_result_msqa} shows the main results on four MSQA datasets. Discriminative models perform better than generative models on the MSQA task, especially on MultiSpanQA-Expand and MAMRC where questions may contain only one answer or no answer. The reason may be that discriminative models are suited for extracting text spans from a given context, whereas generative models are suited for text generation.

After applying the ACC framework, both discriminative models and generative models gain improvements. For instances, the EM F1 score of Tagger (RoBERTa-base) increases from 69.05\% to 72.26\% and the EM F1 score of BART increases from 65.57\% to 67.31\% on MultiSpanQA. For most settings, presicion scores show significant improvements while some recall scores show slight declines, the reason may be that while the classifier successfully identifies some wrong predictions, it also mistakenly classifies some correct predictions as wrong, leading to the exclusion of some correct predictions and thereby lowering the recall scores. In Section~\ref{subsec:performance_of_acc}, we will analyze the classification results of the classifier to verify this point.

We also evaluate the ACC framework with Partial Match P/R/F1 (PM P/R/F1), which considers the overlap between the predictions and gold answers. Results are shown in Appendix~\ref{subsec:pm_results}.

\begin{table}
  \centering
  \small
    \begin{tabular}{l|ccc} 
      \hline
      \multirow{2}{*}{} & \multicolumn{3}{c}{\textbf{MultiSpanQA}} \\ 
       & EM P & EM R & EM F1 \\ 
      \hline
      Tagger BERT & 56.66 & 65.46 & 60.74 \\
      + cls only & 64.90 & 63.98 & 64.44 \\
      + cor only & 62.49 & \textbf{69.11} & 65.63 \\
      + cor \& cls & 67.14 & 67.44 & 67.29 \\
      + binary cls \& cor & 68.58 & 66.56 & 67.56 \\
      + cls \& cor & \textbf{68.52} & 67.05 & \textbf{67.78} \\ 
      \hline
      Tagger RoBERTa & 66.22 & 72.14 & 69.05 \\
      + cls only & 70.54 & 70.58 & 70.56 \\
      + cor only & 68.50 & \textbf{73.09} & 70.72 \\
      + cor \& cls & 71.21 & 71.43 & 71.32 \\
      + binary cls \& cor & 72.45 & 70.94 & 71.68 \\
      + cls \& cor & \textbf{72.39} & 72.12 & \textbf{72.26} \\
      \hline
      \end{tabular}
  \caption{\label{table:ablation_study} Ablation study of ACC framework on the dev set of MultiSpanQA. The best performance is in \textbf{bold}.}
\end{table}

\section{Discussions}

\begin{table}
  \centering
  \small
    \begin{tabular}{l|ccc} 
      \hline
      \multirow{2}{*}{} & \multicolumn{3}{c}{\textbf{MultiSpanQA}} \\ 

      & EM P & EM R & EM F1 \\ 
      \hline
      Tagger BERT & 56.66 & 65.46 & 60.74 \\
      + att cls \& T5 cor & 64.90 & 63.98 & 64.44 \\
      + vanilla cls \& ext cor & \textbf{68.54} & 66.10 & 67.29 \\
      + att cls \& ext cor & 68.52 & \textbf{67.05} & \textbf{67.78} \\ 
      \hline
      Tagger RoBERTa & 66.22 & 72.14 & 69.05 \\
      + att cls \& T5 cor & 70.54 & 70.58 & 70.56 \\
      + vanilla cls \& ext cor & 72.23 & 71.56 & 71.89 \\
      + att cls \& ext cor & \textbf{72.39} & \textbf{72.12} & \textbf{72.26} \\
      \hline
      \end{tabular}
 \caption{\label{table:comparison_different_model} Comparison between diffent combinations of the classifier and the corrector on the validation set of MultiSpanQA. "Att cls" refers to the classifier mentioned in Section~\ref{subsec:cls_model}; "vanilla cls" refers to the classifier without cross-attention layer; "Ext cor" refers to the corrector mentioned in Section~\ref{subsec:cor_model} and "T5 cor" refers to the T5 corrector. The best performance is in \textbf{bold}.}
  \end{table}

\begin{figure*}[t]
  \includegraphics[width=1\linewidth]{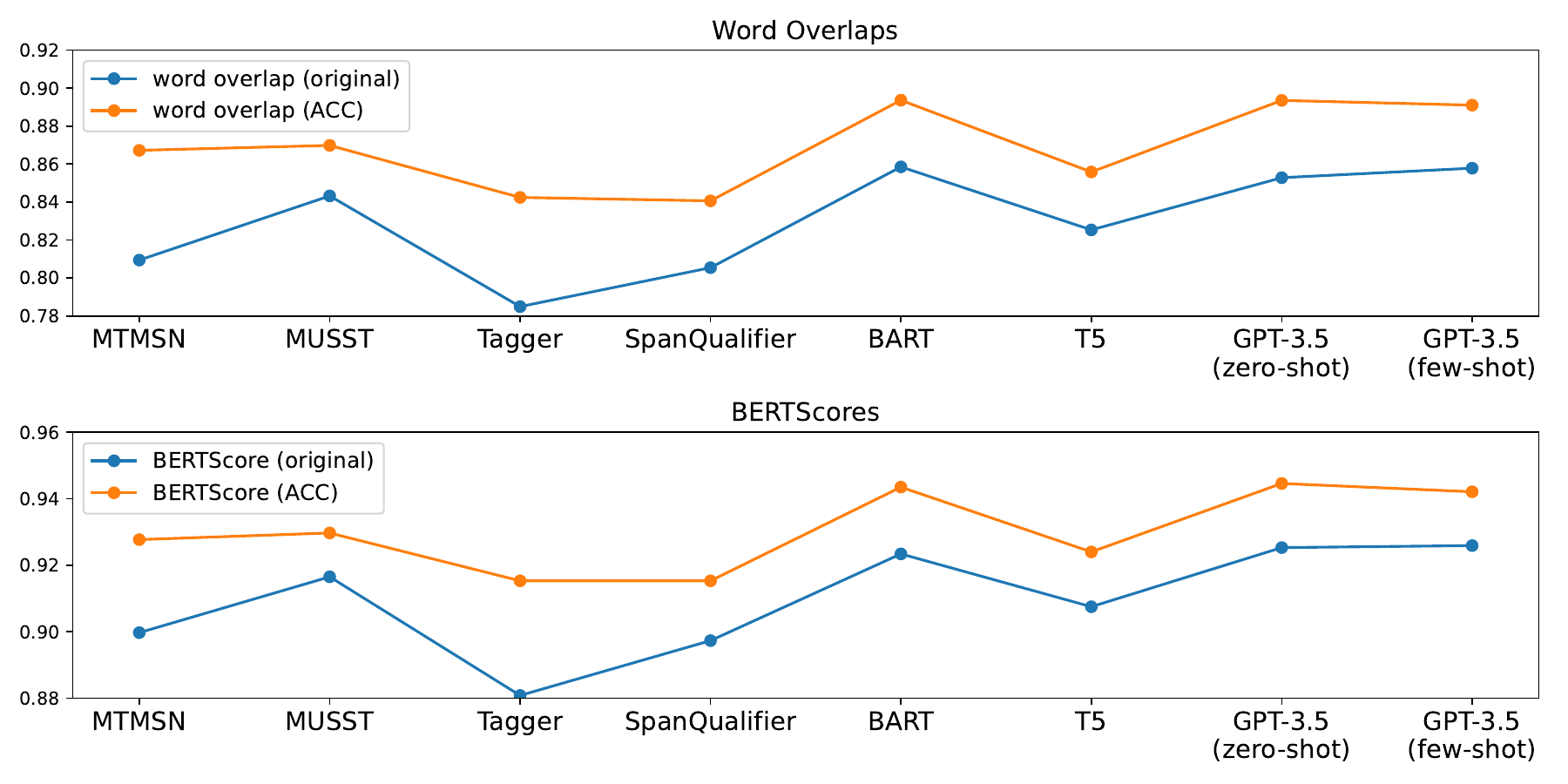}\hfill
  \caption{\textbf{Top:} Average Word Overlap of the predictions. \textbf{Button:} Average BERTScore of the predictions. After applying the ACC framework, both Word Overlap and BERTScore raise, indicating that the ACC framework effectively enhances the quality of the predictions.}
  \label{fig:wo_ss_comparison}
\end{figure*}

\subsection{Ablation Study}

\paragraph{Roles of classifier and corrector.} ACC framework uses the "answer-classify-correct" procedure with the classifier and the corrector. To investigate whether there exists better post-processing procedure, we conduct an ablation study by: 1. only employing the classifier or corrector (cls \textbackslash \ cor only); 2. changing the order of classifier and corrector (cor \& cls); 3. modifying both correct predictions and partially correct predictions (binary cls \& cor).\footnote{For "cls only", we only exclude wrong predictions; for "cor only", we correct all predictions; for "cor \& cls", we first correct all predictions, then classify them and only exclude wrong predictions.}

Table~\ref{table:ablation_study} shows the results of the ablation study on the dev set of MultiSpanQA. The performance of "cls only" and "cor only" lags behind ACC framework, demonstrating the significance of the classifier and corrector. Changing the order between classifier and corrector also shows decline, the reason may be that using corrector first may lead to conceal wrong predictions, thereby the classifier may fail to categorize them as wrong predictions. We also observe that modifying both correct predictions and partially correct predictions does not achieve improvements, demostrating the necessity of distinguishing correct predictions and partially correct predictions and modifying partially correct predictions solely.

\begin{figure*}[t]
  \includegraphics[width=1\linewidth]{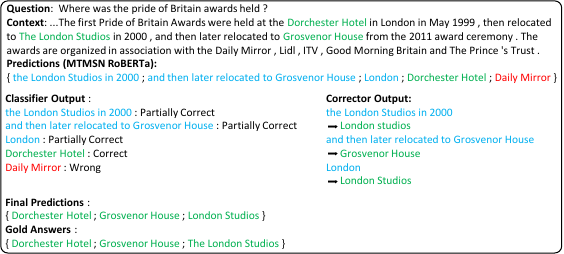}\hfill
  \caption{Case study. The example are selected from the validation set of MultiSpanQA. The correct predictions and gold answers are in \textcolor{green}{green}, the partially correct predictions are in \textcolor{blue}{blue} and the wrong predictions are in \textcolor{red}{red}. Best read in colors.}
  \label{fig:case_study}
\end{figure*}

\paragraph{Comparison with different models.} ACC framework uses a classifier with a cross-attention layer and a corrector based on the pointer model. However, ACC framework can also opt for alternative type of classifiers or correctors. To this end, we replace the classifier and the corrector with other models and compare their performance.\footnote{For the classifier, we replace it with a vanilla classifier where we remove the cross-attention layer; for the corrector, we replace it with T5 \cite{raffel-2020-t5} which outputs texts as the corrected answers.}

Table~\ref{table:comparison_different_model} shows the results of the comparison between different model combinations on the dev set of MultiSpanQA. After replacing the classifier or the corrector, ACC framework shows declines, especially when applying a generative model, ACC framework lag behind other settings. This indicates that the generative models are less capable than traditional pointer models in correcting predictions.

\begin{table}
  \centering
  \resizebox{\linewidth}{!}{
    \begin{tabular}{l|ccc} 
      \hline
      \multicolumn{2}{l}{Tagger BERT} & \multicolumn{1}{l}{} & \multicolumn{1}{l}{} \\ 
      \hline
      label \textbackslash \ pred & wrong & partially & correct \\ 
      \hline
      wrong & \textbf{268 (37.85\%)} & 148 (20.9\%) & 292 (41.24\%) \\
      partially & 16 (6.13\%) & \textbf{98 (37.55\%)} & 147 (56.32\%) \\
      correct & 26 (2.18\%) & 24 (2.01\%) & \textbf{1145 (95.82\%)} \\ 
      \hline
      \multicolumn{2}{l}{Tagger RoBERTa} & \multicolumn{1}{l}{} & \multicolumn{1}{l}{} \\ 
      \hline
      label \textbackslash \ pred & wrong & partially & correct \\ 
      \hline
      wrong & \textbf{135 (27.44\%)} & 105 (21.34\%) & 252 (51.22\%) \\
      partially & 22 (8.63\%) & \textbf{83 (32.55\%)} & 150 (58.82\%) \\
      correct & 27 (2.01\%) & 34 (2.54\%) & \textbf{1280 (95.45\%)} \\
      \hline
      \end{tabular}
}  
\caption{\label{table:cls_model_performance} Accuracy of the classifier on the dev set of MultiSpanQA. The correct classifications of each types are in \textbf{bold}.}
  \end{table}

\subsection{Analysis on the Predictions}
\label{subsec:performance_of_acc}

\paragraph{Accuracy of the classifier.} To analyze the capability of the classifier, we conduct a statistical analysis on its classification results. Table~\ref{table:cls_model_performance} shows the accuracy of the classifier on the dev set of MultiSpanQA. The classifier achieves an high accuracy on the correct predictions (95.82\% for Tagger-BERT and 95.45\% for Tagger-RoBERTa), demonstrating that the ACC framework reserves most correct predictions. On the other hand, the classifier exclude about 1/3 wrong predictions, contributing to the imporvements on EM F1 scores, while the accuracies on the partially true predictions and the wrong predictions can be further improved.

\paragraph{Changes in answers by the corrector.} To analyze the capability of the corrector, we also conduct a statistical analysis on how many prediction has been changed. Table~\ref{table:cor_model_performance} shows the changes of the partially correct predictions on the dev set of MultiSpanQA. The corrector changes 30.77\% of the answers for Tagger-BERT and 27\% for Tagger-RoBERTa, respectively. For Tagger-BERT, 27.47\% of the not-correct predictions are modified to the correct predictions, while 3.3\% of the correct predictions are modified to the not-correct predictions. Furthermore, among all the partially correct predictions derived from the classifier, over 60\% of the incorrect predictions remain incorrect, indicatisng a significant room for improvements.

\begin{table}
  \centering
  \small
    \begin{tabular}{l|cc} 
      \hline
      \multicolumn{2}{l}{Tagger BERT} & \multicolumn{1}{l}{} \\ 
      \hline
      cls \textbackslash \  cls \& cor & incorrect & correct \\ 
      \hline
      incorrect & 172 (63.00\%) & \textbf{75 (27.47\%)} \\
      correct & \textbf{9 (3.30\%)} & 17 (6.23\%) \\ 
      \hline
      \multicolumn{2}{l}{Tagger RoBERTa} & \multicolumn{1}{l}{} \\ 
      \hline
      cls \textbackslash \ cls \& cor & incorrect & correct \\ 
      \hline
      incorrect & 137 (61.43\%) & \textbf{52 (23.32\%)} \\
      correct & \textbf{11 (4.93\%)} & 23 (10.31\%) \\
      \hline
      \end{tabular}
  \caption{\label{table:cor_model_performance} Changes in answers by the corrector on the dev set of MultiSpanQA.}
  \end{table}

\subsection{Analysis on the Quality of the Predictions}

Previous experiment indicates that the ACC framework imporves EM scores. However, the ACC framework may overfit to the annotation boundaries rather than enhancing the quality of the predictions. To this end, we utilize other metrics such as the Word Overlap and BERTScore (mentioned in Section~\ref{subsec:types_of_answers}) and compare the changes of these metrics after applying the ACC framework.

Figure~\ref{fig:wo_ss_comparison} shows the comparison results. Both Word Overlaps and BERTScores raise after applying the ACC framework, with the most significant enhancement in the Tagger where Word Overlap increases by 7\% and BERTScore increases by 4\%. This indicates that the ACC framework enhances the quality of the predictions, rather than overfit to the annotation boundaries. 

We also utilize LLM to evaluate the modified answers, results are shown in Appendix~\ref{subsec:llm_evaluation}.

\subsection{Case Study}

We conduct a case study to demostrate that the ACC framework effectively excludes incorrect predictions and corrects some partially correct predictions. We select a real example where the predictions exactly match the gold answers (i.e. EM F1 = 100\%), shown in Figure~\ref{fig:case_study}. In this example, the MTMSN presents five predictions: "the London Studios in 2000", "and then relocated to Grosvenor House", "London", "Dorchester Hotel" and "Daily Mirror". The classifier identifies “Dorchester Hotel” as a correct prediction and "Daily Mirror" as a wrong prediction. The others three predictions contain irrelevant information or lack specific details, so they are identified as partially correct predictions and modifed by the corrector.\footnote{When calculating EM scores, the article "the" is ignored, so "London Studios" and "The London Studios" are considered as the same prediction.} This example demostrates that our ACC framework effectively enhance the quality of the predictions.

\subsection{Pilot Study with LLM}

ACC framework utilizes a fine-tuned RoBERTa encoder as the backbone. To investigate whether our proposed method works on larger models, we conduct a pilot study by replacing the classifier or corrector with a prompted LLM. The implementation details and prompts are shown in Appendix~\ref{sec:implementation_details_llm}.

Table~\ref*{table:pliot_study_llm} shows the experiment results. After replacing the classifier or the corrector with LLM, the ACC framework still achieves improvements on Tagger-BERT and Tagger-RoBERTa, which proves that our post-processing strategies can be effectively applied to LLM.

\begin{table}
  \centering
  \small
    \begin{tabular}{l|ccc} 
      \hline
      \multirow{2}{*}{} & \multicolumn{3}{c}{\textbf{MultiSpanQA}} \\ 
        & EM P & EM R & EM F1 \\ 
      \hline
      Tagger BERT & 56.66 & 65.46 & 60.74 \\
      +LLM cls \& LLM cor & 68.60 & 63.35 & 65.87 \\
      +LLM cls \& FT cor & \textbf{70.04} & 64.47 & 67.14 \\
      +FT cls \& LLM cor & 67.93 & 66.51 & 67.21 \\
      +FT cls \& FT cor & 68.52 & \textbf{67.05} & \textbf{67.78} \\ 
      \hline
      Tagger RoBERTa & 66.22 & \textbf{72.14} & 69.05 \\
      +LLM cls \& LLM cor & 72.71 & 68.10 & 70.33 \\
      +LLM cls \& FT cor & \textbf{73.69} & 68.97 & 71.25 \\
      +FT cls \& LLM cor & 71.71 & 71.48 & 71.59 \\
      +FT cls \& FT cor & 72.39 & 72.12 & \textbf{72.26} \\
      \hline
      \end{tabular}
  \caption{\label{table:pliot_study_llm} Performance of ACC framework with LLM on the dev set of MultiSpanQA. "LLM cls/cor" refers to classifier/corrector replaced by LLM, and "FT cls/cor" refers to a fine-tuned model. The best performance is in \textbf{bold}.}
\end{table}

\subsection{Model Size and Inference Time}

We analyze the model size and the inference time of the ACC framework. Results and analysis are shown in Appendix~\ref{subsec:model_size_inference_time}.

\section{Related Work}

\subsection{Multi-Span Question Answering}

Recently, a series of MSQA benchmarks \cite{ju-2022-cmqa,li-2022-msqa,yue-2023-mamrc} have been proposed to faclitate research on QA tasks that are closer to real-world scenarios. MSQA tasks require models to extract one or multiple answer spans from a given context. Therefore, traditional SSQA models \cite{seo-2017-bidaf, yu-2018-qanet} are not sufficient to handle multi-span questions.

Existing MSQA methods can be categorized into four categories: (1) pointer-network-based methods. MTMSN \cite{hu-2019-mtmsn} predicts the number of answers, then extracts non-overlapped answer spans; MUSST \cite{yang-2021-musst} uses an autogressive approach to iteratively extract multiple answers. (2) sequence-tagging-based methods. \citet{segal-2020-tase} first convert MSQA task to a sequence-tagging task and utilize BIO tags to mark answer spans; Furthermore, \citet{li-2022-msqa} introduce multi-task learning and achieve better performance. (3) span-enumeration-based methods. SpanQualifier \cite{huang-2023-spanqualifier} utilizes Multi-Layer Perceptron (MLP) to obtain confidence scores for each candidate span and applies a learnable threshold to select answer spans; Similarly, CSS \cite{zhang-2023-css} compares each candidate span with its corresponding question after scoring to obtain answers more similar to the question. (4) LLM-based methods. With the emergence of LLMs like ChatGPT and GPT-4, generative pre-trained language models have been widely applied to various NLP tasks. \citet{zhang-2023-howmany} employ CoT strategies to prompt LLM, and \citet{huang-2023-fbprompt} add negative examples in the few-shot demonstrations.

Existing methods mainly focus on predicting more correct predictions, while the ACC framework takes a post-processing strategy which aims to reduce the number of incorrect predictions. By excluding or modifying incorrect predictions, the ACC framework achieves better performance.

\subsection{Post-Processing Methods}

The post-processing method refers to modifying the original of the model to obtain better predictions. Existing post-processing methods can be categorized into two types: rule-based methods and model-based methods. 

Ruled-based methods typically involve mannually designed rules such as voting to process the outputs from models \cite{campos-2021-biobert, wang-2023-selfconsistency}. On the other hand, model-based methods utilize additional models to modify the hidden states or outputs of the original model, which have been widely applied in Controlled Text Generation (CTG) \cite{yang-klein-2021-fudge, krause-2021-gedi, kim-cho-2023-gta}. In addition to CTG methods, GRACE \cite{khalifa-2023-grace} applies a fine-tuned discriminator to guide language model towards correct multi-step solutions; \citet{ohashi-2023-ppn} utilize a generative model to rewrite the output from a dialogue system and optimize it with Reinforcement Learning (RL) algorithms \cite{stiennon-2020-rl}. 

The work most similar to ours is \cite{gangi-2020-answercorrect}, which utilizes a corrector to modify the outputs of the SSQA model. However, they only focus on partial matches in single-span questions. In constrast, we consider the correctness of multiple predictions in MSQA and additionally employ a classifier to exclude incorrect predictions.

\section{Conclusion}

In this work, we primarily focus on incorrect predictions of the MSQA models. Through a statistical analysis, we observe that models with better performance do not predict less incorrect predictions compared with other models. To this end, we propose ACC framework, which employ a post-processing strategy to exclude wrong predictions and modify partially correct  predictions. Experiments and analysis show that the ACC framework significantly improving the performance by reducing the number of incorrect predictions and obtaining more correct predictions, enhancing the quality of the MSQA predictions.

\section{Limitations and Future Work}

In this work, we categorize incorrect predictions into "partially correct predictions" and "wrong predictions", based on whether the answer should be modified or excluded. However, for "partially correct predictions", there exists more complicated conditions, for example, an incorrect prediction may responses to multiple gold answers. However, the ACC framework can only obtain one modified prediction. In addition, we do not consider the gold answers that MSQA models fail to predict (i.e., "missing predictions"), although the SOTA model still miss 1/3 gold answers. As for future work, we will design more effectively models to handle "partially correct predictions" and "wrong predictions". we will also explore strategies to handle "missing predictions".

\bibliography{custom}

\newpage
\appendix

\begin{table*}
\small
  \centering
  \resizebox{\linewidth}{!}{
    \begin{tabular}{l|cccccccc} 
      \hline
      \multirow{2}{*}{} & \multirow{2}{*}{\#train} & \multirow{2}{*}{\#dev} & \multicolumn{3}{c}{answer number prop.} & \multirow{2}{*}{\begin{tabular}[c]{@{}c@{}}avgerage\\answer number\end{tabular}} & \multirow{2}{*}{\begin{tabular}[c]{@{}c@{}}average\\context length\end{tabular}} & \multirow{2}{*}{\begin{tabular}[c]{@{}c@{}}avgerage\\question length\end{tabular}} \\ 
      \cline{4-6}
       &  &  & $\ge$2 & 1 & 0 &  &  &  \\ 
      \hline
      MultiSpanQA & 5,230 & 658 & 100.0\% & 0.0\% & 0.0\% & 2.89 & 279 & 10 \\
      MultiSpanQA-Expand & 15,690 & 1,959 & 33.4\% & 33.3\% & 33.3\% & 1.30 & 251 & 10 \\
      MAMRC &110,108 & 13,764 & 58.7\% & 41.3\% & 0.0\% & 1.77 & 69 & 10 \\
      MAMRC-Multi & 64,625 & 8,081 & 100.0\% & 0.0\% & 0.0\% & 2.31 & 77 & 10 \\
      \hline
      \end{tabular}
  }
  \caption{\label{table:data_stat} Dataset statistics.}
\end{table*}

\section{Details of Word Overlap and BERTScore}
\label{sec:wo_bs}

\paragraph{Word Overlap.} Assuming that a prediction $p_i$ contains $k$ words $\{p_{i1},p_{i2},...,p_{ik}\}$ and a gold answer $a_j$ contains $l$ words $\{a_{j1},a_{j2},...,a_{jl}\}$, the Word Overlap is defined as Equation~\ref{eq:wo}:

\begin{equation}
  \label{eq:wo}
  WO(p_i,a_j)=\frac{|p_i \cap a_j|}{max(k,l)}
\end{equation}
where $|A|$ denotes the number of element in the set $A$.

\paragraph{BERTScore\cite{zhang-2020-bertscore}.} BERTScore primarily calculates the semantic similarity between the candidate text and the reference text using cosine similarity. Given candicate text $X$ with $m$ tokens $\{x_1,x_2,...,x_m\}$ and reference text $Y$ with $n$ tokens $\{y_1,y_2,...,y_n\}$, BERTScore first computes the cosine similarity $s_{ij}$ between each pair of token vectors $x_i$ and $y_j$. Then, it maximizes the similarity score using a greedy matching approach to calculate the precision score $P(X,Y)$ and the recall score $R(X,Y)$. Finally, it computes the harmonic mean of these two scores (i.e., the F1 score) to obtain the final BERTScore. The above process can be represented by Equation~\ref{eq:bs1}-\ref{eq:bs4}. \footnote{For simplicity, Equation~\ref{eq:bs1}-\ref{eq:bs4} do not consider inverse document frequency (idf) weighting or scaling of $R(c, r)$. For more details of BERTScore, please refer to \cite{zhang-2020-bertscore}}

\begin{equation}
\label{eq:bs1}
  s_{ij} = \frac{H^{x_i} \cdot H^{y_j}}{||H^{x_i}||\ ||H^{y_j}||}
\end{equation}
\begin{equation}
  \label{eq:bs2}
  P(X,Y)=\frac{1}{m}\sum^{m}_{i=1} \max_{j} s_{ij}
\end{equation}
\begin{equation}
  \label{eq:bs3}
  R(X,Y)=\frac{1}{m}\sum^{m}_{i=1} \max_{j} s_{ij}
\end{equation}
\begin{equation}
  \label{eq:bs4}
  BS(X,Y)=2 \cdot \frac{P(X, Y) \cdot R(X, Y)}{P(X, Y) + R(X, Y)}
\end{equation}
where $H^{x_i}$ and $H^{y_j}$ are the representations of $x_i$ and $y_j$ from a Pre-trained Language Model, $||a||$ denotes the length of the vector $a$.

\section{More Details of Experimental Setup}

\subsection{Datasets}
\label{subsec:details_of_datasets}

\paragraph{MultiSpanQA and MultiSpanQA-Expand \cite{li-2022-msqa}} : MultiSpanQA and MultiSpanQA-Expand focus on multi-span questions. The raw questions and contexts are extracted from the Natural Question dataset \cite{kwiatkowski-2019-nq}. MultiSpanQA only contains multi-span questions, while MultiSpanQA-Expand contains both multi-span questions, single-span questions and unanswerable questions.

\paragraph{MAMRC and MAMRC-Multi \cite{yue-2023-mamrc}} : MAMRC is a large-scale dataset containing over 100,000 questions, including both multi-span questions and single-span questions. To investigate the performance on the multi-span questions, we select multi-span questions from MAMRC and obtain MAMRC-Multi. 

Since the official test sets of these four datasets are not public, we report the performance on validation sets. Some statistics about the four datasets are shown in Table~\ref{table:data_stat}.

\subsection{MSQA models}
\label{subsec:details_of_baselines}

\paragraph{MTMSN \cite{hu-2019-mtmsn}}: MTMSN adds a classification head to predict the number of answers. During the inference stage, for each question, MUSST first obtains top-20 predictions and predict answer number $K$, then applies Non-Maximum Sampling algorithm \cite{rosenfold-1971-nms} to extract $K$ non-overlapped spans.

\paragraph{MUSST \cite{yang-2021-musst}}: MUSST adds $m$ linear layer to predict the start position and end position of $m$ spans, where $m$ is the maximum answer number in the training dataset. During the inference stage, MUSST applies an autogressive decoding strategy, where in each iteration MUSST masks out predicted spans and chooses top-1 predictions. The iterative process terminates when the model predicts no more answers or the number of predictions reaches the maximum answer number.

\paragraph{Tagger}: Following the implementation of \cite{li-2022-msqa}, we utilize BIO tags to label each token in context: the first token of the answer is labeled with "B", the other tokens of the answer are labeled with "I" and the tokens not in an answer are labeled with "O".

\paragraph{SpanQualifier \cite{huang-2023-spanqualifier}}: SpanQualifier enumerates all possible answer spans and obtains their corresponding confidence scores as correct predictions, then utilizes a learnable threshold to select the correct prediction spans, achieving state-of-the-art performance on MultiSpanQA-Expand dataset.

\paragraph{BART \cite{lewis-2020-bart} and T5 \cite{raffel-2020-t5}}: Both BART and T5 are pre-trained models with encoder-decoder architecture, which are commonly used in text generation tasks. In this work, we use the delimiter "\#" to concatenate multiple answers and fine-tune the models in a sequence-to-sequence form.

\paragraph{GPT-3.5}: GPT-3.5 is one of the most commonly used LLMs today and can be accessed via API \footnote{\url{https://platform.openai.com/.}}. In our work, we select gpt-3.5-turbo-0120 for our experiments and set up both zero-shot and few-shot prompts. The zero-shot prompt contains only a basic description of the MSQA task, while the few-shot template includes several demonstrations. Specifically, we apply In-Context Learning (ICL) \cite{brown-2020-icl} and utilize a BM25 retriever \cite{robertson-1994-bm25} to select the demonstrations which is similar to the questions. The prompts are shown in Table~\ref{table:baseline_prompt}.

\begin{table*}
\small
  \centering
  \resizebox{\linewidth}{!}{
\begin{tabular}{l|ccc|ccc|ccc|ccc} 
\hline
\multirow{2}{*}{} & \multicolumn{3}{c|}{\textbf{MultiSpanQA}} & \multicolumn{3}{c|}{\textbf{MultiSpanQA-Expand}} & \multicolumn{3}{c|}{\textbf{MAMRC}} & \multicolumn{3}{c}{\textbf{MAMRC\_Multi}} \\
 & PM P & PM R & PM F1 & PM P & PM R & PM F1 & PM P & PM R & PM F1 & PM P & PM R & PM F1 \\ 
\hline
\multicolumn{13}{c}{Discriminative Models (BERT-base)} \\ 
\hline
MTMSN & 69.97 & 79.23 & 74.30 & 73.29 & 73.46 & 73.37 & 84.59 & 89.62 & 87.03 & 84.68 & 89.97 & 87.25 \\
+ACC & \textbf{81.10} & 66.77 & 73.24 & \textbf{77.20} & 67.04 & 71.76 & \textbf{88.45} & 85.86 & \textbf{87.13} & \textbf{90.74} & 85.68 & \textbf{88.14} \\ 
\hline
MUSST & 76.39 & 68.76 & 72.38 & 77.79 & 70.99 & 74.22 & 87.25 & 88.25 & 87.74 & 87.75 & 87.69 & 87.72 \\
+ACC & \textbf{81.25} & 65.68 & \textbf{72.64} & \textbf{78.36} & 68.65 & 73.17 & \textbf{88.68} & 85.46 & 87.04 & \textbf{90.93} & 84.61 & 87.66 \\ 
\hline
Tagger & 78.27 & 77.92 & 78.09 & 70.60 & 65.75 & 68.05 & 88.81 & 89.05 & 88.92 & 88.23 & 84.98 & 86.57 \\
+ACC & \textbf{83.30} & 77.29 & \textbf{80.19} & \textbf{74.06} & \textbf{66.64} & \textbf{70.14} & \textbf{89.07} & 87.13 & 88.09 & \textbf{90.85} & 83.54 & \textbf{87.04} \\ 
\hline
SpanQualifier & 81.17 & 79.70 & 80.43 & 74.01 & 76.73 & 75.34 & 87.75 & 90.94 & 89.31 & 87.55 & 91.90 & 89.67 \\
+ACC & \textbf{84.26} & 77.70 & \textbf{80.84} & \textbf{76.20} & 75.15 & \textbf{75.67} & \textbf{88.83} & 87.94 & 88.38 & \textbf{90.78} & 88.37 & 89.56 \\ 
\hline
\multicolumn{13}{c}{Discriminative Models (RoBERTa-base)} \\ 
\hline
MTMSN & 77.57 & 82.29 & 79.86 & 76.36 & 76.80 & 76.58 & 85.77 & 89.72 & 87.70 & 85.15 & 90.18 & 87.60 \\
+ACC & \textbf{85.65} & 72.12 & 78.30 & \textbf{78.88} & 69.93 & 74.14 & \textbf{88.74} & 86.21 & 87.46 & \textbf{90.45} & 86.08 & \textbf{88.21} \\ 
\hline
MUSST & 83.44 & 75.72 & 79.39 & 80.22 & 73.36 & 76.63 & 88.64 & 88.44 & 88.54 & 88.65 & 86.64 & 87.63 \\
+ACC & \textbf{85.41} & 73.24 & 78.86 & 79.99 & 70.83 & 75.13 & \textbf{89.42} & 85.95 & 87.65 & \textbf{91.17} & 83.89 & 87.38 \\ 
\hline
Tagger & 83.97 & 83.92 & 83.94 & 77.91 & 75.43 & 76.64 & 90.09 & 90.22 & 90.15 & 88.07 & 85.90 & 86.98 \\
+ACC & \textbf{86.60} & 82.67 & \textbf{84.59} & \textbf{79.43} & 74.62 & \textbf{76.95} & 89.92 & 89.01 & 89.46 & \textbf{90.81} & 84.20 & \textbf{87.38} \\ 
\hline
SpanQualifier & 83.85 & 83.17 & 83.50 & 76.77 & 78.62 & 77.65 & 89.82 & 88.19 & 89.00 & 87.27 & 92.14 & 89.63 \\
+ACC & \textbf{86.39} & 81.27 & \textbf{83.74} & \textbf{78.69} & 76.67 & \textbf{77.66} & 89.34 & \textbf{88.98} & \textbf{89.16} & \textbf{90.49} & 88.82 & \textbf{89.65} \\ 
\hline
\multicolumn{13}{c}{Generative Models} \\ 
\hline
BART-base & 85.76 & 74.85 & 79.94 & 76.35 & 66.40 & 71.02 & 87.99 & 84.25 & 86.08 & 88.44 & 82.83 & 85.55 \\
+ACC & \textbf{88.22} & 72.85 & 79.81 & \textbf{76.58} & 64.26 & 69.88 & 87.99 & 84.25 & 86.08 & 88.18 & 80.91 & 84.39 \\ 
\hline
T5-base & 86.70 & 79.48 & 82.93 & 81.06 & 74.81 & 77.81 & 87.52 & 88.45 & 87.98 & 87.41 & 85.56 & 86.47 \\
+ACC & \textbf{88.15} & 76.88 & 82.13 & \textbf{81.22} & 72.22 & 76.46 & 86.69 & 85.99 & 86.34 & 87.40 & 83.22 & 85.26 \\ 
\hline
GPT3.5 (Zeroshot) & 85.70 & 79.64 & 82.56 & 57.62 & 66.03 & 61.54 & 61.10 & 73.70 & 66.81 & 72.27 & 77.71 & 74.89 \\
+ACC & \textbf{89.64} & 74.73 & 81.51 & \textbf{64.38} & 64.07 & \textbf{64.23} & \textbf{67.19} & 69.23 & \textbf{68.19} & \textbf{78.24} & 74.55 & \textbf{76.35} \\ 
\hline
GPT3.5 (Fewshot) & 88.19 & 81.28 & 84.59 & 59.67 & 70.81 & 64.76 & 72.47 & 85.98 & 78.65 & 79.27 & 87.04 & 82.98 \\
+ACC & \textbf{90.76} & 78.23 & 84.03 & \textbf{67.07} & 68.27 & \textbf{67.67} & \textbf{77.55} & 81.59 & \textbf{79.52} & \textbf{83.54} & 83.97 & \textbf{83.75} \\
\hline
\end{tabular}
  }
  \caption{\label{table:pm_score_msqa} PM scores on four MSQA datasets.}
\end{table*}

\subsection{Implementation Details}
\label{subsec:implementation_details}

To determine the hyper parameters $\alpha$ and $\beta$, we analyze the Word Overlap and BERTScore of the sampled data, shown in Table~\ref{table:wo_bs_distribution}. For the middle 60\% of sampled data, the Word Overlaps range from 0 to 0.25, and the BERTScore range from 0.36 to 0.62. Based on this, we set $\alpha$ to 0.25 and $\beta$ to 0.6.

When sampling training data for ACC framework, we set split number $K=3$, which means in each iteration, we use two-thirds of the training data for training and sample the predictions on the remaining data. for the classifier, we maintain a balanced ratio of 1:1:1 among the three answer categories for the classifier, and for the corrector, we added examples that require no modifications and maintained a ratio of 2:1 between examples requiring modifications and examples requiring no modifications, considering that corrector may not necessarily modifies all the input predictions.

During training stage of classifier and corrector, for MultiSpanQA and MultiSpanQA-Expand, we set $learning\_rate=3e^{-5}$, $batch\_size=48$, $epochs=10$ and $max\_length=512$; For MAMRC and MAMRC-Multi, we set $learning\_rate=3e^{-5}$, $batch\_size=96$, $epochs=5$ and $max\_length=256$. We choose the best classifier and corrector on our sliver-labeled validation sets. All the baselines were trained with three different seeds and we report the mean results. We perform our experiments on a single Tesla V-100 GPU(32GB).

\section{Additional Experiments and Discussions}

\subsection{Partial Match Results}
\label{subsec:pm_results}

The Partial Match results are shown in Table~\ref{table:pm_score_msqa}. While EM F1 scores show significant improvements after applying the ACC framework, PM F1 scores achieve less improvements and even decline in some cases. The main reason may be that PM scores consider the overlaps between predictions and gold answers, as a result, incorrect predictions may contribute to PM F1 score (i.e., $EM\ F1 = 0, PM\ F1>0$). However, such predictions are not desired and may be excluded by the ACC framework, limiting the improvements in PM F1 scores.

\subsection{Evaluation of the predictions with LLM}
\label{subsec:llm_evaluation}

\begin{table}
  \centering
  \small
    \begin{tabular}{l|cc} 
    \hline
     & Word Overlap & BERTScore \\ 
    \hline
    Min. & 0.00 & 0.00 \\
    Max. & 0.96 & 1.00 \\
    Avg. & 0.11 & 0.49 \\
    Mid. 60\% Range & (0.00,0.25) & (0.36,0.62) \\
    \hline
\end{tabular}
  \caption{\label{table:wo_bs_distribution} The distribution infomation of the sampled data on Word Overlap and BERTScore (metric we use to define partially correct prediction and wrong prediction). "Min." refers to the minimum value, "Max." refers to the maximum value, "Avg." refers to the average value, and "Mid. 60\% Range" refers to the range of the middle 60\% of the data.}
\end{table}

We utilize LLM to evaluate whether the predictions are closer to the gold answers after applying the ACC framework. For each dataset, we collect the predictions modified by the ACC framework and randomly sample 500 pairs (including original prediction, new prediction and gold answers for each pair) for evaluation. We mannully label four pairs as the few-shot demostractions for GPT-3.5. The prompts are shown in Table~\ref{table:llm_evaluate_prompt}.

The evaluation results are shown in Table~\ref{table:llm_evaluation}. Across the four datasets, the LLM consider approximately 80\% of the new preictions to be closer to the gold answers. This indicates that the ACC framework improves the quality of the predictions.

\begin{table}
  \centering
  \small
  \begin{tabular}{lll} 
    \hline
    \multicolumn{1}{l}{Datasets} & \multicolumn{1}{c}{original} & \multicolumn{1}{c}{new} \\ 
    \hline
    MultiSpanQA & 89 (17.8\%) & 411 (82.2\%) \\
    MultiSpanQA-Expand & 98 (19.6\%) & 402 (80.4\%) \\
    MAMRC & 98 (19.6\%) & 402 (80.4\%) \\
    MAMRC-Multi & 92 (18.4\%) & 408 (81.6\%) \\
    \hline
    \end{tabular}
  \caption{\label{table:llm_evaluation} LLM evaluation on which prediction is closer to the gold answer, where "original" indicates that GPT-3.5 judge the original prediction to be closer and "new" indicates that GPT-3.5 judge the prediction modified by the ACC framework to be closer. }
\end{table}

\subsection{Model Size and Inference Time}
\label{subsec:model_size_inference_time}

We compare model sizes between MSQA models and the ACC framework, shown in Table~\ref{table:model_parameters}. The ACC framework improves the performance of baselines without applying large-size models, avoiding consuming excessive computational resources.

\begin{figure}[t]
  \includegraphics[width=1\linewidth]{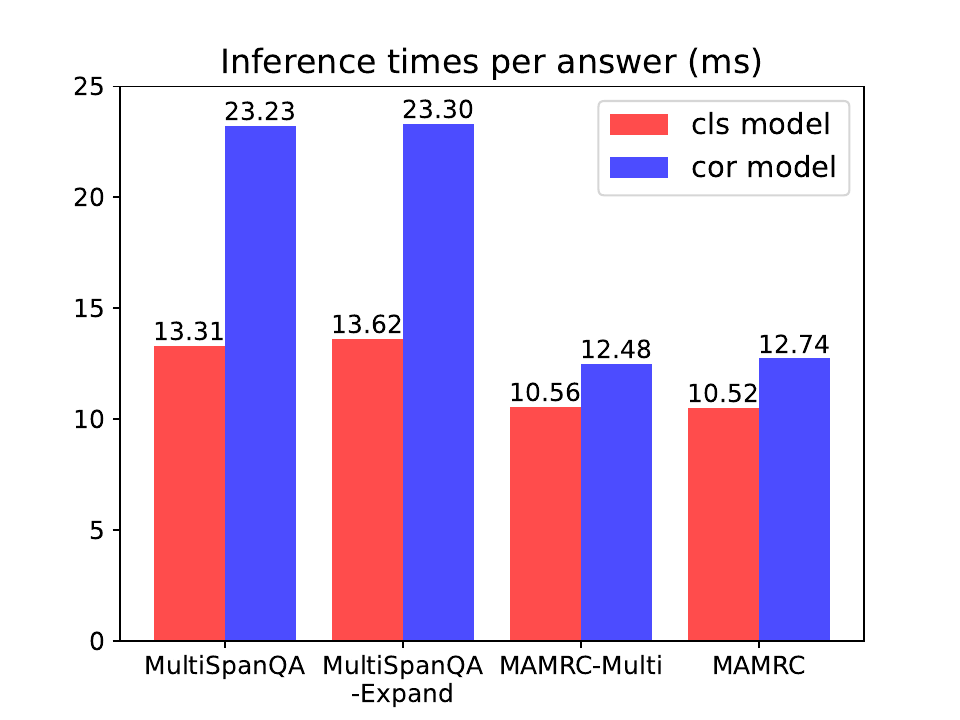}\hfill
  \caption{Inference times on four datasets.}
  \label{fig:inference_time}
\end{figure}

We also analyze inference times of the ACC framework, shown in Figure~\ref{fig:inference_time}. The results demostrate that the ACC framework is time-effective, especially when the input length is short (we set $max\_length=256$ for MAMRC and MAMRC-Multi and we set $max\_length=512$ for MultiSpanQA and MultiSpanQA-Expand).

\begin{table}
  \centering
  \small
    \begin{tabular}{lcc} 
      \hline
      \textbf{model} & \textbf{BERT-base} & \textbf{RoBERTa-base} \\ 
      \hline
      MTMSN & 110M & 125M \\
      MUSST & 110M & 125M \\
      Tagger & 109M & 125M \\
      SpanQualifier & 115M & 131M \\ 
      \hline
      classifier & - & 128M \\
      corrector & - & 124M \\
      \hline
      \end{tabular}
  \caption{\label{table:model_parameters} Model sizes of baselines model, the classifier and the corrector.}
\end{table}

\subsection{Implementation details of pilot study with LLM}
\label{sec:implementation_details_llm}

We use OpenAI's official API \footnote{\url{https://platform.openai.com/.}} and select the model gpt-3.5-turbo-0120 for our pilot study. Due to the poor performance in zero-show settings, we apply In-Context Learning (ICL) \cite{brown-2020-icl} and utilize a BM25 retriever \cite{robertson-1994-bm25} to select the demonstrations which is similar to the questions. When replacing the classifier, we select one demonstration for each answer type; when replacing the corrector, we select two demostractions for answers requiring modification and requiring no modification. The prompts are shown in Table~\ref{table:pliot_study_prompt}.

\begin{table*}
  \centering
  \resizebox{\linewidth}{!}{
    \begin{tabular}{l} 
      \hline
      \textbf{Instruction:} \\ 
      \hline
      \begin{tabularx}{\textwidth}{X}For this task, we will provide you a passage and a question. The question contains one or multiple answers and these answers are in the passage. You should first read the given passage and question, then extract answer spans from the passage and use "\#" to split each answer spans, i.e. answer1\#answer2\#answer3.You should output your answer in a json format like "\{"answer":"your\_answer"\}", DO NOT include any explanations in your responses. \\\end{tabularx} \\
      \hline
      \textbf{Demostractions (for few-shot setting):} \\ 
      \hline
      \begin{tabularx}{\textwidth}{X}Example 1:\\Passage: ...\\Question: ...\\Answer: ...\\... \\\end{tabularx} \\
      \hline
      \textbf{Query:} \\ 
      \hline
      \begin{tabularx}{\textwidth}{X}Query:\\Passage: ...\\Question: ...\\Answer: \\\end{tabularx} \\
      \hline
    \end{tabular}
  }
  \caption{\label{table:baseline_prompt} Prompts for zero-shot LLM reader and few-shot LLM reader.}
\end{table*}

\begin{table*}
  \centering
  \resizebox{\linewidth}{!}{
    \begin{tabular}{l} 
      \hline
      \textbf{Instruction:} \\ 
      \hline
      \begin{tabularx}{\textwidth}{X}For this task, we will provide you with the gold answer to a question, the original prediction from our AI model, and a new prediction modified by another AI model. The question is from a QA dataset. You need to determine which prediction, the original or the new, is more accurate and closer to the gold answer. \\\end{tabularx} \\
      \hline
      \textbf{Demostractions (for few-shot setting):} \\ 
      \hline
      \begin{tabularx}{\textwidth}{X}Example 1:\\Original Prediction: Billy\\New Prediction: Billy Jorl\\Gold: Billy Joel\\Answer: new \\\end{tabularx} \\
      \hline
      \textbf{Query:} \\ 
      \hline
      \begin{tabularx}{\textwidth}{X}Original Prediction: ...\\New Prediction: ...\\Gold: ...\\Answer: \\\end{tabularx} \\
      \hline
    \end{tabular}
  }
  \caption{\label{table:llm_evaluate_prompt} Prompts for the evaluation on the predictions with LLM.}
\end{table*}

\begin{table*}
  \centering
  \resizebox{\linewidth}{!}{
    \begin{tabular}{l} 
      \hline
      \textbf{cls model prompt:} \\ 
      \hline
      \begin{tabularx}{\textwidth}{X}For this task, we will provide you a passage and a question. The question contains one or multiple answers and these answers are in the passage. We will also provide you a candidate answer from our AI model. You should read the passage, the question and classify the candidate answer into one of three classes: "correct prediction", "partially correct prediction" and "wrong prediction". Correct prediction refers to a completely correct prediction; Partially correct prediction refers to a prediction that is basically correct but still requires some modifications. Wrong prediction refers to a prediction that is completely incorrect and should be excluded. You should output your answer in a json format like "\{\{"answer":"your\_answer"\}\}", DO NOT include any explanations in your responses.\\Example1: \\Passage: ... \\Question: ...\\Candidate Answer:...\\Output: \{"answer":"correct prediction"\}\\...\\Query:\\Passage: ...\\Question: ...\\Candidate Answer: ...\\Output:\end{tabularx} \\ 
      \hline
      \textbf{cor model prompt:} \\ 
      \hline
      \begin{tabularx}{\textwidth}{X}For this task, we will provide you a passage and a question. The question contains one or multiple answers and these answers are in the passage. We will also provide a candidate answer that our AI model believes needs some modifications. You should read the passage, the question and judge whether the candidate answer requires modifications. If no modifications are needed, you should output the candidate answer as is. Otherwise, you should modify it by adding or deleting some words, and the modified prediction must be a part of the passage and similar to the original candidate answer. You should output your answer in a json format like "\{\{"answer":"your\_answer"\}\}", DO NOT include any explanations in your responses.\\Example1: \\Passage: ... \\Question: ...\\Original Answer: ...\\Output: \{"answer":"xxx"\}\\...\\Query:\\Passage: ...\\Question: ...\\Candidate Answer: ...\\Output:\end{tabularx} \\
      \hline
      \end{tabular}
  }
  \caption{\label{table:pliot_study_prompt} Prompts for pilot study with LLM}
\end{table*}

\end{document}